\documentclass[letterpaper,10pt,conference]{ieeeconf}

\IEEEoverridecommandlockouts
\usepackage{amsmath,amssymb}
\usepackage{booktabs}
\usepackage{multirow}
\usepackage{placeins}
\usepackage{graphicx}
\usepackage{algorithm}
\usepackage{algorithmicx}
\usepackage{algpseudocode}
\usepackage{xcolor}
\usepackage{microtype}
\usepackage{tikz}
\usetikzlibrary{arrows.meta,fit,positioning}
\usepackage{xurl}
\usepackage[hidelinks,bookmarks=false]{hyperref}

\title{\fontsize{18}{22}\selectfont\bfseries
AdaHVLA: Adaptive Harnesses for Long-Horizon\\
Vision-Language-Action Execution}

\author{%
\authorblockN{\fontsize{12}{14.4}\selectfont\bfseries
Junyi Tang\textsuperscript{1}, Jie Peng\textsuperscript{2}, Zezhen Ding\textsuperscript{3}, Yuan Shen\textsuperscript{4}, Tianlong Chen\textsuperscript{1,*}}
\authorblockA{\textsuperscript{1}UNC\quad
\textsuperscript{2}USTC\quad \textsuperscript{3}HKUST\quad \textsuperscript{4}CUHK\quad
\textsuperscript{*}Corresponding Author\\[3pt]
Code: \href{https://github.com/Haaareally/AdaHVLA-Adaptive_Harness_VLA}{\nolinkurl{github.com/Haaareally/AdaHVLA-Adaptive_Harness_VLA}}}}

\begin{document}

\bstctlcite{BSTcontrol}

\maketitle
\thispagestyle{empty}
\pagestyle{empty}

\begin{abstract}

Vision-language-action (VLA) models offer strong local control and instruction following but often struggle with long-horizon tasks requiring persistent memory and planning. Task harnesses
provide persistent context for agent reasoning by retaining task
history and tracking progress across execution stages.
To bring these complementary capabilities together, we introduce AdaHVLA, an adaptive harness that refines code-based coordination policies through robot experience to better align agent reasoning and memory with VLA execution. Its decoupled multiagent adaptation process separates evidence analysis, harness revision, and behavioral assessment into distinct working contexts, using testable coordination hypotheses to guide revisions and subsequent rollouts to assess their predicted effects. A stateful revision graph links execution evidence, hypotheses, revisions, and observed effects, preserving alternative harnesses and adaptation memory to guide refinement across repeated attempts and continued adaptation across tasks and environments. In simulation, AdaHVLA raises mean test success on NaVILA-LH from 22.5\% to as high as 57.5\% and improves manipulation test success across three VLA backbones by up to 30.8 percentage points over the initial harness. Real-world deployment further illustrates how the adapted policies support stable execution across task stages.

\end{abstract}

\section{Introduction}
\label{sec:introduction}

Real-world robot tasks rarely end with a single action. A quadruped
may need to inspect several rooms in sequence, while a robotic arm
must leave objects in configurations suitable for later assembly
stages. Vision-language-action (VLA) models have made substantial
progress in visual grounding and language-conditioned execution
for various embodied tasks~\cite{kim2025openvla,black2025pi0, cheng2025navila}. Yet a locally appropriate action
does not ensure progress toward the full task. The robot must retain
relevant history, track what remains to be done, and adjust its
behavior when execution departs from the plan~\cite{
shi2026memoryvla,wang2026progressthink,huang2023innermonologue}.
Each stage also establishes conditions that allow the next to
succeed~\cite{fan2025longvla}. Reliable long-horizon execution
therefore requires coordination beyond local action prediction.

\begin{figure}[!t]
    \vspace{1.5mm}
    \centering
    \includegraphics[width=\linewidth]{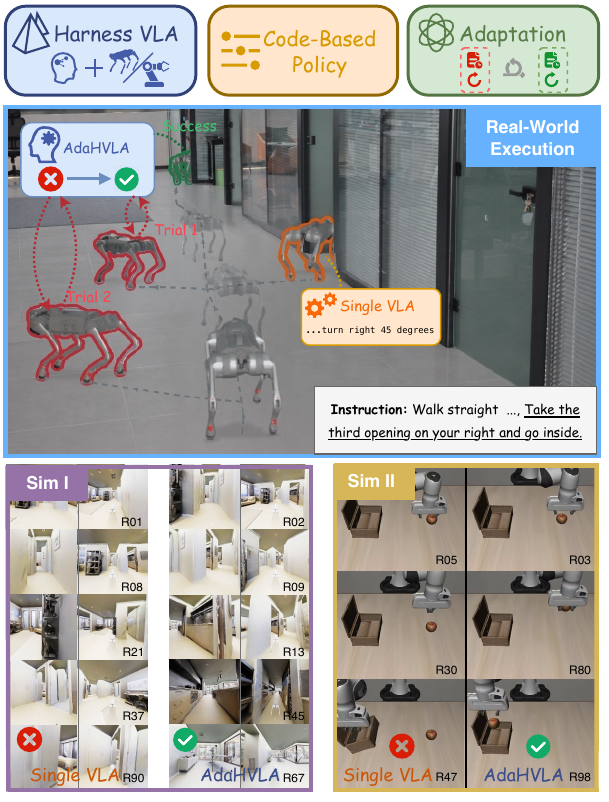}
    \vspace{-8mm}
    \caption{
    AdaHVLA deployed in both simulated and real-world environments, adapting across diverse tasks and robotic platforms.
    }
    \label{fig:intro_show}
    \vspace{-3mm}
\end{figure}

Agents built on large language models (LLMs) and vision-language
models (VLMs) complement local execution with broader task
reasoning. They can interpret observations and execution history
in light of an overall goal, revising plans as the task
unfolds~\cite{huang2023innermonologue,huang2023voxposer}.
Code as Policies translates instructions into executable programs
that compose perception and control
interfaces~\cite{liang2023codeaspolicies}.
Hierarchical VLA systems take a complementary route, connecting
task reasoning to learned local execution through explicit
guidance interfaces~\cite{shi2025hirobot,li2025hamster}.
We adopt this division of responsibility: an executable task
harness maintains context across stages and governs when and how
agent reasoning guides the VLA, while the VLA carries out local
instructions using visual observations.

Effective cooperation, however, involves more than choosing the
next skill. Planning over a predefined skill repertoire can ground
decisions in the robot's available
capabilities~\cite{ichter2023saycan}, but skill selection alone does
not determine what history to retain, when to advance a stage, or
how long to maintain corrective guidance. For example, arriving at
a doorway is not equivalent to passing through it; advancing the
task too early can redirect subsequent instructions away from an
unfinished passage. Coordination rules may therefore need to
change as layouts, executor responses, and intermediate task
states vary, even when the initial design works under familiar
conditions. This motivates adapting the coordination policy
itself, rather than only revising the plan under fixed rules.

To this end, we introduce AdaHVLA (Fig.~\ref{fig:intro_show}), an adaptive
harness VLA that refines code-based coordination policies through robot
experience to better align agent reasoning and memory with VLA
execution. These editable policies determine which observations and history the VLA receives, when to advance to the next task stage,
and how to respond to deviations and decide whether the task
is complete. This flexibility allows coordination to
be tailored to the executor and environment, but a plausible code
revision need not produce its intended behavioral effect.
Execution outcomes depend on interacting perception, reasoning,
and control processes, making code validity alone an insufficient
measure of improvement~\cite{fu2026capx}.
AdaHVLA therefore revises policies between robot rollouts through
a multiagent process with distinct working contexts for evidence
analysis, harness revision, and behavioral assessment.
Testable coordination hypotheses specify which mechanism should
change and what behavior should follow. Subsequent rollouts
provide evidence for assessing these predictions, connecting
proposed revisions to their effects on robot execution.

To make adaptation cumulative, a revision graph preserves
alternative harnesses and links execution evidence, hypotheses,
revisions, and observed effects. A useful revision need not solve
an entire task immediately: preventing an early stop may expose
a later difficulty that warrants further refinement. The graph
retains this evidence to guide candidate selection and further
evaluation, allowing both successful and unsuccessful attempts to
inform subsequent revisions. Within a task, repeated episodes
provide further checks on a candidate's effects and support
continued policy refinement. Across tasks and environments, the
revised harness and accumulated adaptation memory guide later
adaptation. The code shapes future execution, while the records
preserve what was changed, why, and with what effect.

Our contributions are as follows:
\begin{itemize}
    \item We introduce AdaHVLA, an adaptive harness that couples
    agent reasoning and memory with VLA execution through
    code-based coordination policies. It translates robot
    experience into policy refinement, enabling editable policies
    to evolve with tasks and execution conditions.

    \item We develop a multiagent adaptation algorithm that
    separates evidence analysis, harness revision, and behavioral
    assessment. Testable coordination hypotheses and a revision
    graph connect proposed changes to execution evidence,
    guiding candidate selection and further evaluation within
    and across tasks.

    \item We evaluate AdaHVLA on simulated quadruped navigation
    and robot manipulation. Across different models, navigation test success rises from 22.5\% with the initial
    harness to 31.7--57.5\%. Adaptation improves manipulation
    test success across three VLA backbones on both long-horizon
    levels, while real-world deployment further illustrates
    stable execution across task stages.
\end{itemize}
\section{Related Work}
\label{sec:related}

\subsection{Long-Horizon Robot Execution}

Long-horizon navigation and manipulation benchmarks emphasize
skill composition, subtask dependencies, and sustained task
progress~\cite{song2025longhorizon,zhang2025vlabench}.
Recent models address these demands through phase-aware training
for subtask compatibility~\cite{fan2025longvla}, temporal
memory~\cite{shi2026memoryvla}, and semantic progress
estimation~\cite{wang2026progressthink}.
Explicit planning offers a complementary way to organize
execution. SayCan grounds skill selection in language and robotic
affordances~\cite{ichter2023saycan}, while Inner Monologue
incorporates environmental feedback into ongoing
planning~\cite{huang2023innermonologue}.
Code as Policies generates programs that compose perception
interfaces and control primitives~\cite{liang2023codeaspolicies};
other approaches express spatial objectives through value maps
or relational keypoint constraints for motion
planning~\cite{huang2023voxposer,huang2025rekep}.

Hierarchical architectures connect task reasoning to learned
control through intermediate representations. Hi Robot uses
language subgoals to guide a local VLA from complex instructions
and feedback~\cite{shi2025hirobot}, whereas HAMSTER supplies
coarse two-dimensional end-effector paths to a separate control
policy~\cite{li2025hamster}.
These interfaces can be supplemented with memory across
reasoning and acting layers~\cite{li2025longhorizonvla} and
executable checks around an existing policy, as in
Code-as-Monitor~\cite{zhou2025codeasmonitor}.
AdaHVLA builds on these architectures by adapting the code-based
coordination policies between agent reasoning and VLA execution
through robot experience.

\begin{figure*}[!t]
    \centering
    \includegraphics[width=\textwidth]{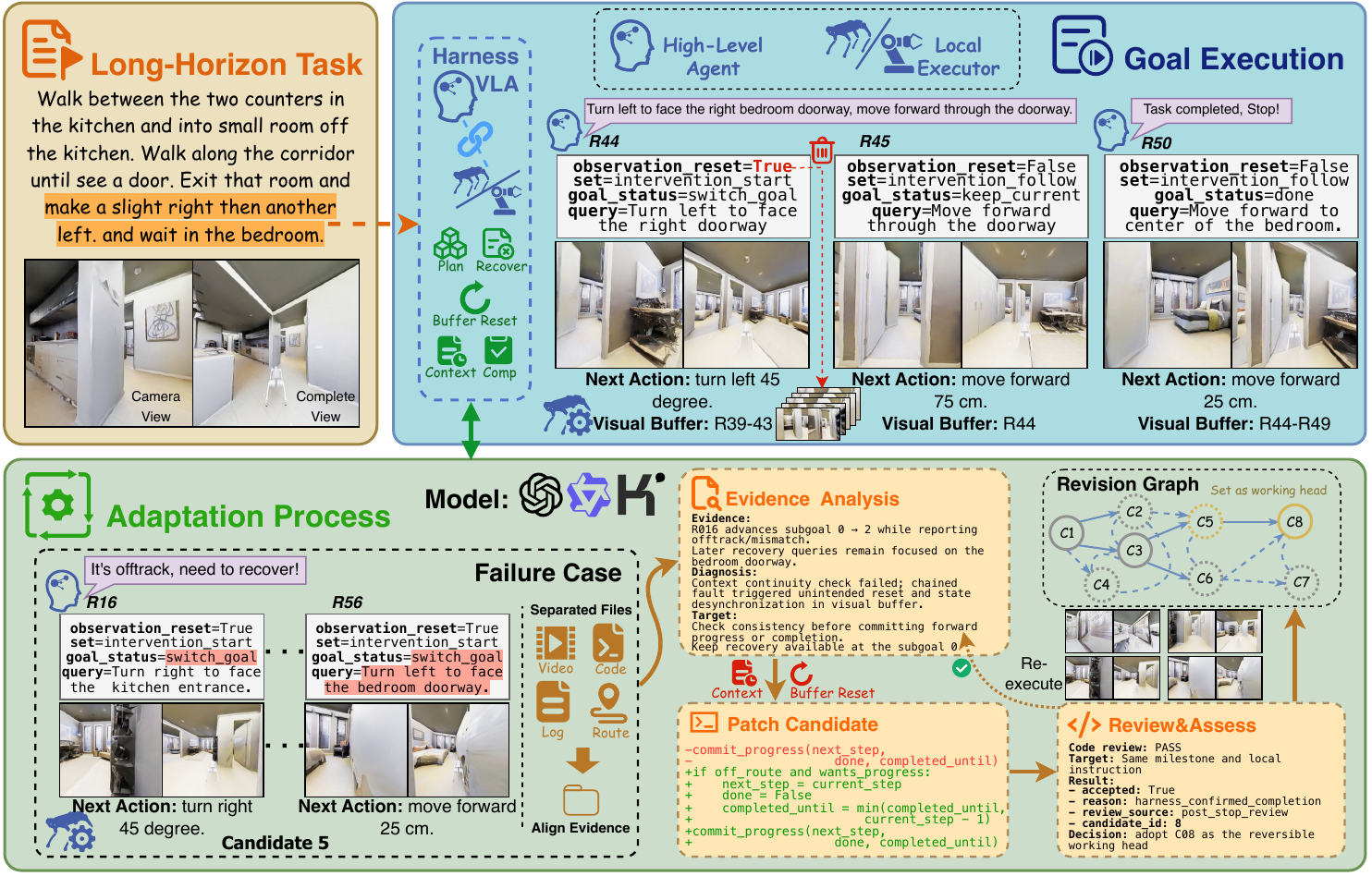}
    \vspace{-7mm}
    \caption{Overview of AdaHVLA. During execution (top), code-based
        policies connect task reasoning and history
        to VLA execution. During adaptation (bottom), execution
        evidence informs testable coordination hypotheses and code
        revisions, whose effects are assessed in new rollouts. A revision graph preserves
        candidate policies and linked evidence for continued adaptation.
        Each $R_i$ denotes the request ID of a harness--VLA interaction.}
    \label{fig:framework}
    \vspace{-3mm}
\end{figure*}

\vspace{1.2mm}
\subsection{Embodied Adaptation and Harness Optimization}

Execution feedback can be retained as memory or used to revise an
agent's executable logic. Reflexion incorporates feedback into
linguistic memory~\cite{shinn2023reflexion}, while automated agent
design searches for improved programs using evaluated
candidates~\cite{hu2025adas}.
Related methods optimize the organization of agent computation:
AFlow searches executable workflows~\cite{zhang2025aflow}, and
GPTSwarm optimizes agent components and their
connections~\cite{zhuge2024gptswarm}.
These approaches provide mechanisms for exploring alternative
implementations, but embodied adaptation also requires validating the closed-loop behavior induced by those implementations~\cite{agia2025unpacking,zhou2025autoeval}.

In robotics, execution traces support failure diagnosis and
program refinement, while validated repairs can be retained
as reusable procedural knowledge~\cite{fu2026capx,lu2026aspire}.
These advances strengthen robot execution, yet successful
re-execution alone does not establish whether a revision
produces its intended behavioral effect or how that evidence
should guide further adaptation. AdaHVLA addresses these
questions for code-based coordination policies between agent
reasoning and VLA execution. Its multiagent process separates
evidence analysis, harness revision, and behavioral assessment
into distinct working contexts: testable coordination
hypotheses guide revisions, and subsequent rollouts assess
their predicted effects. A revision graph links execution
evidence, hypotheses, candidate policies, and observed effects
as adaptation memory, allowing both successful and unsuccessful
attempts to inform the selection of candidate revisions and continued refinement
within and across tasks and environments.
\section{Method}
\label{sec:method}

\subsection{Overview}
\label{sec:overview}

AdaHVLA adapts the code-based coordination policies connecting
agentic reasoning and memory to VLA execution
(Fig.~\ref{fig:framework}). During execution, the harness uses
observations and task history to guide the VLA and update task
state. Between rollouts, a multiagent process turns execution
evidence into testable coordination hypotheses, code revisions,
and behavioral assessments. A revision graph links these
records to alternative policies, supporting refinement over
repeated attempts within a task. The revised harness and
adaptation memory persist across tasks and environments,
providing a basis for subsequent revisions.

\subsection{Problem Formulation}
\label{sec:formulation}

Let $\xi=(x,\mathcal E,s_0)$ denote a task instance with instruction
$x$, environment $\mathcal E$, and initial robot and scene state
$s_0$. At decision boundary $t$, the harness coordinates VLA $\pi_\theta$
using interaction history $\mathcal H_t=(x,o_{\leq t},a_{<t})$
and runtime state $m_t$:
\begin{equation}
    \begin{aligned}
        (q_t,\kappa_t,m_{t+1},z_t)
        &\sim H_c(\cdot\mid\mathcal H_t,m_t),\\
        a_t &\sim \pi_\theta(\cdot\mid q_t,\kappa_t),
        \qquad z_t=0.
    \end{aligned}
    \label{eq:harness_policy}
\end{equation}
Here $H_c$ is the policy induced by source revision $c$ and
reasoning-agent calls; $o_t$ is the observation and $a_t$ a
VLA command or action sequence passed to the controller.
The outputs specify a VLA instruction $q_t$, observation
context $\kappa_t$, updated state $m_{t+1}$, and task termination
flag $z_t\in\{0,1\}$. An episode is one rollout: $c$ remains
fixed, $m_t$ is initialized at its start, and $z_t=1$ ends
execution.

For existing revisions $\mathcal C$, adaptation seeks
\begin{equation}
    \max_{c\in\mathcal C}\;
    \mathbb E_{\xi\sim\mathcal D,\,
    \tau\sim P_c(\cdot\mid\xi)}
    [R(\tau;\xi)],
    \label{eq:adaptation_objective}
\end{equation}
where $\mathcal D$ is the task distribution, $P_c$ the induced
rollout distribution, and $R(\tau;\xi)\in\{0,1\}$ the benchmark
success indicator. Revisions use a finite adaptation set,
with the robot configuration and VLA parameters unchanged.
Behavioral assessments guide this search for task completion.

\subsection{Harness VLA Policy}
\label{sec:runtime_coordination}

AdaHVLA exposes the
interface between agentic reasoning and VLA execution as a set of
editable, executable coordination policies. These policies, illustrated in Fig.~\ref{fig:framework}, determine
what state persists across calls, what instruction and context reach
the VLA, when task progress is committed, and when recovery or
completion logic takes control. 

The harness maintains runtime state $m_t$ and constructs the
context in which reasoning-agent calls interpret
observations and task progress. The resulting decisions are used
to form the instruction $q_t$ and observation context $\kappa_t$
passed to the VLA. The VLA produces a command or action sequence
for the controller, whose execution yields observations for
subsequent coordination. Thus, the harness determines what the
executor should address and which context it receives; the VLA
supplies the local action behavior.

\textit{Task progress and guidance.}
Ordered sub-tasks and goals specify observable criteria for progress.
The policy determines when to advance the active goal
and whether to retain the original instruction or provide a
local goal as $q_t$. Advancement depends on the state left
by previous actions, including whether new-goal conditions have been established.

\textit{Context and visual refresh.}
The harness selects observations, progress records, and ongoing
corrections for reasoning, while controlling visual refresh
at the executor interface. These uses of historical observations can require
different retention rules: an earlier visual buffer may remain useful
for route reasoning after a turn but become stale for local
execution.

\textit{Recovery and completion.}
The policy determines when corrective guidance begins,
persists, and ends. In Fig.~\ref{fig:framework}, guidance
continues from a corrective turn to forward motion through
the intended doorway. Completion is checked against remaining
goals, so a local stopping signal need not terminate
the full task. Revisions change how these decisions interact
across VLA calls, rather than the VLA's action model.

A fixed source revision $c$ does not imply a fixed sequence of
instructions. Within a rollout, new observations and updates to
$m_t$ can change the active goal, retained context, and recovery
decisions under the same code. The same task instruction can
therefore lead to different local requests as execution unfolds,
without regenerating source during the episode. Between rollouts,
adaptation changes the code governing these decisions. It leaves
VLA parameters unchanged, so improvements are sought through
how existing reasoning and execution capabilities are coordinated.

\subsection{Policy Revision and Assessment}
\label{sec:policy_revision}
\label{sec:behavioral_assessment}

Each revision specifies an expected behavioral effect before
execution. Evidence analysis, harness revision, and behavioral
assessment use distinct agent contexts, exchanging structured
records rather than a shared conversation.

\textit{From evidence to revision.}
For each rollout, camera observations, logs, and
trajectory segments are aligned by request ID and linked to
the harness source revision
(Fig.~\ref{fig:framework}). An analysis agent examines this
evidence, recording observed events separately from explanations.
Each testable coordination hypothesis identifies a suspected
mechanism, supporting evidence, an expected effect, and an observable
indicator. For example, advancing a goal while off-route motivates retaining
it to keep guidance focused on the unfinished passage.

A revision agent inspects the source and related attempts,
then proposes $c'=c\oplus\Delta$, where $\Delta$ is a code patch
and $\oplus$ denotes its application. Source inspection may
alter the explanation; the hypothesis and predicted effect
underlying the patch are recorded before a new rollout.
A separate source and interface review establishes eligibility
for execution; subsequent rollouts assess behavioral improvement.

\textit{Assessing the predicted effect.}
A separate agent compares parent and revised rollouts on the
same task instance in a fresh context, assessing the targeted
effect, overall behavior, and final success separately.
Task outcomes also depend on perception and control, which
can obscure the contribution of a code
revision~\cite{fu2026capx}. We therefore examine whether the
modified mechanism was triggered and produced its predicted effect,
rather than crediting task success alone.

Let $\hat u$ denote whether the modified mechanism was triggered,
and $\hat v$ whether the trajectories had already diverged
before that trigger. Both are assessed as true ($1$), false
($0$), or undetermined ($\bot$). We define the attribution gate
\begin{equation}
    \chi=\mathbf{1}\{\hat u=1\land\hat v=0\},
    \label{eq:attribution_gate}
\end{equation}
where $\mathbf{1}\{\cdot\}$ is the indicator function.
Only comparisons with $\chi=1$ can update hypothesis support:
targeted improvements strengthen support, whereas unchanged
or degraded effects weaken it when the candidate is
discontinued. Otherwise, support is retained, while all
outcomes remain available for subsequent refinement.
The gate restricts attribution from observed behavior;
repeatability requires further rollouts rather than inference
from a single successful episode.

\subsection{Revision Graph and Memory}
\label{sec:revision_graph}
\label{sec:revision_search}

Following only the latest revision can entangle unresolved
coordination issues with subsequent changes and lead to
repeated, unproductive attempts. Motivated by branching
exploration and the reuse of intermediate
results~\cite{zhang2026dgm,besta2024graphofthoughts},
AdaHVLA maintains a revision graph $\mathcal G$ linking code
versions, hypotheses, rollout evidence, and behavioral
assessments. Related attempts remain retrievable across branches, allowing further
work to draw on earlier revisions and their outcomes.

\begin{table*}[!t]
    \centering
    \vspace*{5.5pt}
    \caption{Published model benchmarks and NaVILA-LH success rates (\%).}
    \label{tab:navigation_results}
    \small
    \setlength{\tabcolsep}{2pt}
    \renewcommand{\arraystretch}{1.08}
    \begin{tabular*}{\textwidth}{@{\extracolsep{\fill}}lcccccccc@{}}
        \toprule
        \multirow{2}{*}{\shortstack[l]{System /\\adaptation model}}
        & \multicolumn{2}{c}{Published benchmarks}
        & \multirow{2}{*}{\shortstack{Adaptation\\SR}}
        & \multicolumn{4}{c}{Test Success Rate}
        & Overall SR \\
        \cmidrule(lr){2-3}\cmidrule(lr){5-8}\cmidrule(l){9-9}
        & TB 2.1 & MMMU-Pro &
        & Repeat 1 & Repeat 2 & Repeat 3 & Mean
        & Mean $\pm$ SD \\
        \midrule
        Single VLA & N/A & N/A & $30.0$
        & --- & --- & --- & $2.5$ & --- \\
        Initial Harness VLA & N/A & N/A & $36.7$
        & --- & --- & --- & $22.5$ & --- \\
        \midrule
        GPT-5.6-sol Max~\cite{openai2026gpt56} & $88.8$ & $83.0$ & $\mathbf{73.3}$
        & $62.5$ & $52.5$ & $57.5$ & $\mathbf{57.5}$
        & $\mathbf{60.7\pm6.1}$ \\
        GPT-5.6-terra Max~\cite{openai2026gpt56} & $87.4$ & $80.7$ & $63.3$
        & $47.5$ & $60.0$ & $37.5$ & $48.3$
        & $51.3\pm7.0$ \\
        Kimi K3~\cite{moonshot2026kimik3} & $88.3$ & $81.6$ & $56.7$
        & $32.5$ & $55.0$ & $42.5$ & $43.3$
        & $46.0\pm11.1$ \\
        Qwen3.7-Max~\cite{qwen2026qwen37max} & NR & NR & $53.3$
        & $52.5$ & $27.5$ & $40.0$ & $40.0$
        & $42.7\pm12.1$ \\
        GPT-5.6-luna Max~\cite{openai2026gpt56} & $84.7$ & $78.4$ & $46.7$
        & $25.0$ & $47.5$ & $32.5$ & $35.0$
        & $37.3\pm7.6$ \\
        Qwen3.6-Max~\cite{qwen2026qwen36max} & NR & NR & $43.3$
        & $42.5$ & $20.0$ & $32.5$ & $31.7$
        & $34.0\pm10.0$ \\
        \bottomrule
    \end{tabular*}
    \par\vspace{0.5mm}
    \begin{minipage}{\textwidth}
        \footnotesize
        \textit{Notes.} TB 2.1 denotes Terminal-Bench 2.1;
        MMMU-Pro scores use no tools. Published scores follow
        their source evaluation protocols, including different
        coding harnesses~\cite{openai2026gpt56,moonshot2026kimik3}.
        NR: no verified score for the listed configuration;
        N/A: no adaptation model; ``---'': statistic not reported.
    \end{minipage}

    \vspace{-4.5mm}
\end{table*}

\textit{Policy exploration within a task.}
The graph informs which candidate to continue, which hypothesis
to reconsider, and where another rollout is needed.
Candidates are prioritized by task success, followed by
targeted and overall behavioral effects. Let $W$ denote the
maximum number of active candidate leaves and $D$ the maximum
revision depth within the current task focus. Within the
adaptation budget, a revised candidate is eligible for
extension when its assessment identifies a targeted improvement
or a newly exposed failure and capacity remains.
This decision is distinct from updating hypothesis support:
an incomplete task can still provide a useful direction for
further refinement. Inactive candidates and their evidence
remain available, so an alternative line of work can resume
without reconstructing earlier attempts. Repeated rollouts
and checks on previously successful adaptation tasks examine
whether improvements recur.

\textit{Adaptation memory across tasks.}
Relevant agent context is condensed into long-term adaptation memory,
retaining coordination issues, tested hypotheses, attempted
changes, observed effects, and the conditions under which
they were evaluated, with links to the supporting evidence.
The selected harness and retained agent context persist across
task switches. Retrieved summaries guide new hypotheses and
revisions, while runtime state and local exploration depth
reset. Revision contexts are rebuilt for unrelated assignments,
and each behavioral assessment starts afresh.
Code preserves the adapted coordination policies; adaptation
memory preserves the evidence and unresolved questions needed
to refine them under new task and environmental conditions.

The final harness is selected using adaptation evidence alone.
Its source remains fixed on held-out tasks, while online
reasoning and runtime state updates continue.

\section{Experiments}
\label{sec:experiments}

Long-horizon execution places different coordination demands on
robots navigating through scenes and manipulating objects whose
states constrain later actions. We evaluate whether AdaHVLA can turn execution experience into
effective coordination on held-out instances across task demands,
robots, and model configurations.

\subsection{Long-Horizon Benchmarks}

\noindent\textbf{Quadruped navigation.}
We construct \emph{NaVILA-LH} from 50 NaVILA-Bench
instances~\cite{cheng2025navila},
randomly assigning 10 to adaptation and 40 to testing.
Tasks require traversing passages, following turns, and stopping
at an intended destination. Completing the instruction requires
long-horizon execution as the robot moves through the scene,
rather than finishing each local command in isolation.

\smallskip
\noindent\textbf{Robot manipulation.}
We use the VLA-Arena Long Horizon suite~\cite{zhang2026vlaarena},
with 50 instances per level and a 10/40 adaptation/test split
at each level. L1 combines independent atomic skills; L2
introduces longer workflows and dependencies between stages,
increasing task complexity.
These tasks extend the evaluation from tracking progress along
a route to coordinating actions whose resulting object states
determine whether subsequent goals can be achieved.

\begin{table}[!b]
    \vspace{-3.5mm}
    \centering
    \caption{Manipulation success rates (\%) across VLA backbones.}
    \label{tab:manipulation_results}
    \small
    \setlength{\tabcolsep}{2pt}
    \renewcommand{\arraystretch}{1.00}
    \setlength{\aboverulesep}{0.25ex}
    \setlength{\belowrulesep}{0.35ex}
    \begin{tabular*}{\columnwidth}{@{\extracolsep{\fill}}lccccc@{}}
        \toprule
        \multirow{2}{*}{\shortstack{VLA\\backbone}}
        & \multirow{2}{*}{\shortstack{Params\\(B)}}
        & \multirow{2}{*}{\shortstack{Adaptation\\SR}}
        & \multicolumn{3}{c}{Test Success Rate} \\
        \cmidrule(l){4-6}
        & & & Single & Initial & Adapted \\
        \midrule
        \multicolumn{6}{@{}l}{\textit{L1}} \\
        SmolVLA~\cite{shukor2025smolvla}
        & $0.45$ & $40.0$ & $7.5$ & $11.7$ & $25.8$ \\
        OpenVLA-OFT~\cite{kim2025oft}
        & $7.5$ & $56.7$ & $15.0$ & $20.8$ & $40.8$ \\
        $\pi_{0.5}$~\cite{black2025pi05}
        & $\sim 3.3$ & $\mathbf{70.0}$ & $20.0$ & $25.0$
        & $\mathbf{55.8}$ \\
        \midrule
        \multicolumn{6}{@{}l}{\textit{L2}} \\
        SmolVLA~\cite{shukor2025smolvla}
        & $0.45$ & $16.7$ & $1.7$ & $2.5$ & $10.8$ \\
        OpenVLA-OFT~\cite{kim2025oft}
        & $7.5$ & $33.3$ & $4.2$ & $7.5$ & $23.3$ \\
        $\pi_{0.5}$~\cite{black2025pi05}
        & $\sim 3.3$ & $\mathbf{50.0}$ & $6.7$ & $10.0$
        & $\mathbf{38.3}$ \\
        \bottomrule
    \end{tabular*}
\end{table}

\begin{figure*}[!t]
    \centering
    \includegraphics[width=0.96\textwidth]{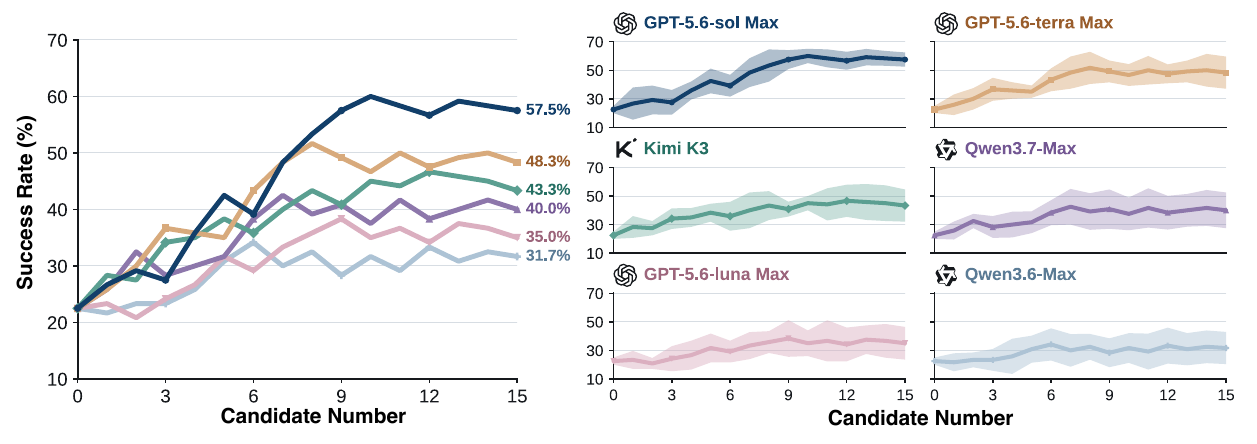}
    \vspace{-3mm}
    \caption{
    NaVILA-LH test success by candidate. Each point gives the indexed
    candidate's mean test SR over three adaptation restarts.
    The left panel compares six adaptation models; the right panels
    add sample SD. Candidate 0 is Initial Harness VLA, followed by
    candidates 1--15. The runtime reasoning agent and
    NaVILA~\cite{cheng2025navila} remain unchanged. Test outcomes
    do not guide revisions or candidate selection.
    }
    \label{fig:comparison}
    \vspace{-3mm}
\end{figure*}

\subsection{Setup}

\noindent\textbf{Simulation and control.}
Navigation uses a Unitree Go2 in Matterport3D scenes through
Isaac Lab and Isaac Sim, with a pretrained locomotion controller
converting NaVILA~\cite{cheng2025navila} commands into robot
motion. Manipulation uses a Panda robot in RoboSuite with
MuJoCo through VLA-Arena~\cite{zhang2026vlaarena}. The reference
environment provides RGB observations and seven-dimensional
actions; observation inputs follow each VLA adapter to support execution.

\smallskip
\noindent\textbf{Comparisons and adaptation.}
We compare long-horizon instruction execution (\emph{Single VLA}),
the non-adapted coordination policy (\emph{Initial Harness VLA}),
and the policy adapted by AdaHVLA. The initial harness retains
online reasoning and state updates but undergoes no code revision.
Both harnesses use fixed source code during held-out testing.
Navigation uses the six adaptation configurations in
Table~\ref{tab:navigation_results}; manipulation uses
GPT-5.6-sol Max~\cite{openai2026gpt56}.
All harnesses use Qwen3.7-Flash~\cite{alibaba2026qwen37flash} as the runtime reasoning agent.
Navigation uses NaVILA~\cite{cheng2025navila} as its executor;
manipulation uses three VLA backbones:
SmolVLA~\cite{shukor2025smolvla},
OpenVLA-OFT~\cite{kim2025oft}, and
$\pi_{0.5}$~\cite{black2025pi05}.

For each configuration, we restart adaptation three times from
the initial harness on the same task split. AdaHVLA autonomously
evaluates up to sixteen candidates per run (0--15, including the
initial harness). VLA parameters and controllers remain fixed.
Runtime ablations require no readaptation.

\smallskip
\noindent\textbf{External capability references.}
Published Terminal-Bench 2.1 and MMMU-Pro scores provide coding
and visual reasoning references~\cite{openai2026gpt56,moonshot2026kimik3}.
They follow their source protocols and do not guide adaptation.

\smallskip
\noindent\textbf{Evaluation.}
Success rate (SR) follows the benchmark completion criterion.
Adaptation instances guide revision and selection; test instances
are held out. Rates average three runs on the same tasks:
adaptation restarts for adapted policies and repeated evaluation
for fixed baselines. Means thus summarize 30 adaptation or
120 test evaluations, not distinct tasks or the total adaptation
rollout budget. Overall SR pools both splits within each run.
Table~\ref{tab:navigation_results} lists adapted-harness test
repeats and baseline means; the adaptation column in
Table~\ref{tab:manipulation_results} refers to the adapted harness.
Means are rounded to one decimal place after averaging; SD
denotes sample standard deviation across runs. Test SR measures
reuse beyond adaptation instances.

\subsection{Quadruped Navigation}

Table~\ref{tab:navigation_results} separates the benefit of
introducing coordination from that of refining it through experience.
Initial Harness VLA already raises mean test SR from 2.5\% to
22.5\%, a 20.0-point gain from coordinating the fixed executor.
AdaHVLA then reaches 31.7--57.5\% across six adaptation models,
adding 9.2--35.0 percentage points over this initial policy.
With the runtime agent and VLA unchanged, these held-out gains
show that execution experience can improve how reasoning and
local control cooperate beyond the initial coordination rules.

The 25.8-point spread among adapted endpoints suggests that
improvement depends on the adaptation model. Published coding and
visual reasoning scores follow different protocols and do not fully
track navigation performance across model families~\cite{openai2026gpt56,moonshot2026kimik3},
so they serve as capability references. Figure~\ref{fig:comparison}
traces revisions. As in Section~\ref{sec:revision_graph}, candidates
for refinement are prioritized by adaptation-task success, then
targeted and overall behavioral effects.
Tables~\ref{tab:navigation_results} and~\ref{tab:manipulation_results}
report final harnesses selected using adaptation evidence alone.

\subsection{Robot Manipulation}

Manipulation tests the same adaptation architecture with a
different robot and execution interface
(Table~\ref{tab:manipulation_results}). Initial Harness VLA
improves on Single VLA in all six backbone--level combinations,
and adaptation improves further in every setting, reaching
25.8--55.8\% on L1 and 10.8--38.3\% on L2.
With $\pi_{0.5}$~\cite{black2025pi05}, the initial harness raises
L1 test SR from 20.0\% to 25.0\%; adaptation adds another
30.8 points to reach 55.8\%. On L2, adaptation raises SR from
10.0\% to 38.3\%, a comparable 28.3-point gain despite longer
workflows with dependencies between stages.
The repeated improvement over an already useful initial policy
in Tables~\ref{tab:navigation_results} and~\ref{tab:manipulation_results}
supports a common adaptation architecture that specializes
coordination for each robot--executor configuration while
keeping the parameters of the underlying VLA unchanged.

The benefit spans different model sizes, although absolute
performance does not increase monotonically with parameter
count: $\pi_{0.5}$~\cite{black2025pi05} outperforms the larger
OpenVLA-OFT~\cite{kim2025oft} on both levels. This comparison
does not isolate model size; the gains within each backbone
show the value of adapting coordination around an existing executor.
Across all three backbones, absolute gains and final SRs remain
lower on L2 than on L1. Both workflow length and stage dependencies
vary between levels, so their individual effects cannot be
separated by the present comparison.

\section{Discussion}
\label{sec:discussion}

The initial harness already improves execution. Adaptation
adds value by using behavioral evidence and separately managed
contexts to refine coordination for each robot.

\begin{figure}[!b]
    \vspace{-2mm}
    \centering
    \includegraphics[width=\columnwidth]{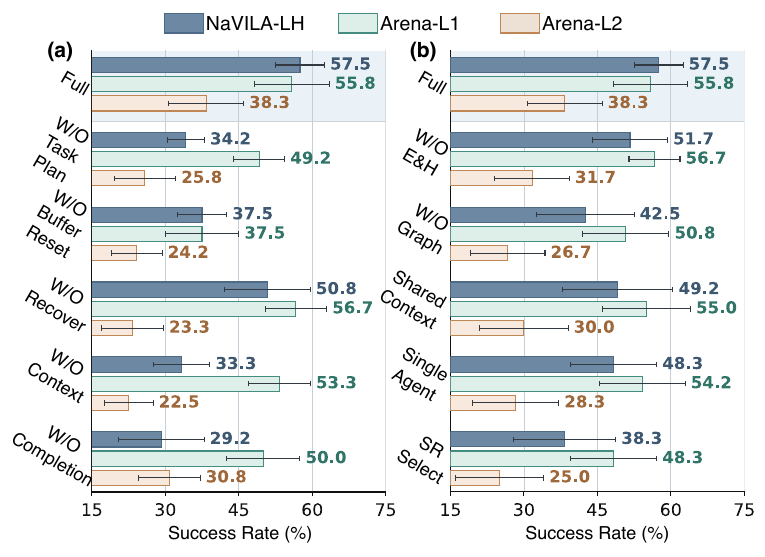}
    \vspace{-6.6mm}
    \caption{
    Ablations of (a) execution policies after adaptation and
    (b) the adaptation process from a common initial harness.
    Manipulation uses $\pi_{0.5}$~\cite{black2025pi05}.
    Error bars denote sample SD over three runs.
    }
    \label{fig:ablation}
\end{figure}

\begin{figure*}[!t]
    \centering
    \includegraphics[width=\textwidth]{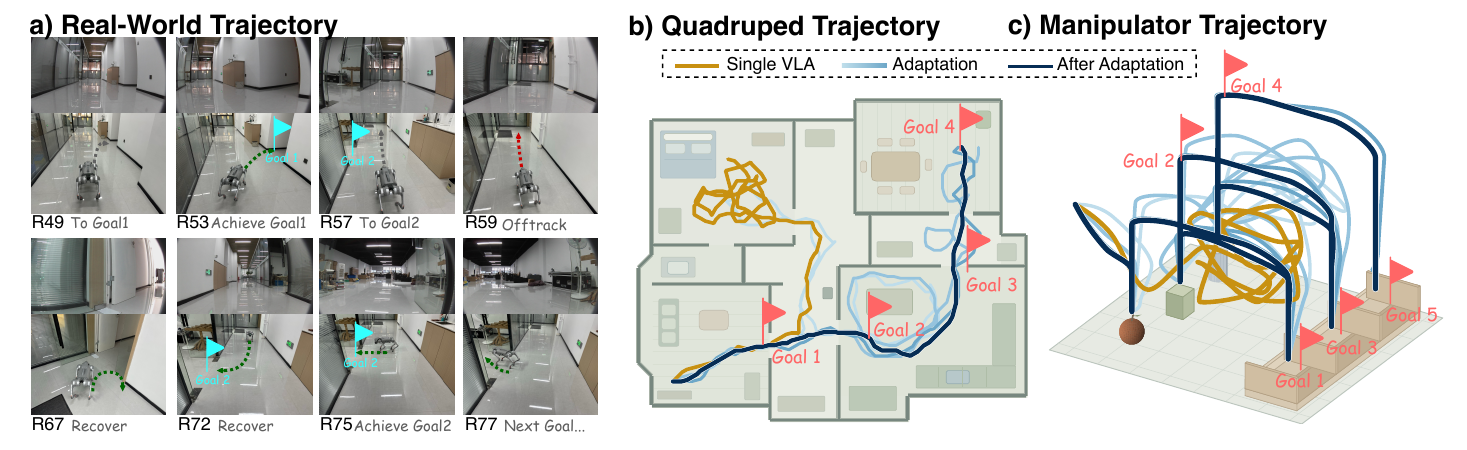}
    \vspace{-7mm}
    \caption{
    Execution across embodiments and environments.
    (a) Real-world Unitree Go2 deployment: red and green arrows
    indicate deviation and goal-directed motion.
    Recorded trajectories for (b) simulated navigation and
    (c) manipulation compare Single VLA with harness policies
    during and after adaptation.
    }
    \label{fig:trajectory}
    \vspace{-3.3mm}
\end{figure*}

\subsection{Cumulative Policy Improvement}
\label{sec:policy_iteration}

As adaptation moves across tasks, retained agent context and
linked revision evidence let later candidates build on earlier
attempts (Section~\ref{sec:revision_graph}). The overall gains in
Fig.~\ref{fig:comparison} are consistent with this cumulative
refinement, although progress varies across adaptation models
and later candidates can regress. Preserving earlier revisions
therefore gives the process alternatives to refine when a
recent change proves less useful.

Continued refinement requires both reusable evidence and
appropriate working context. Section~\ref{sec:ablation} examines
these choices through graph and context ablations, alongside
behavioral versus success-only candidate selection.

\subsection{Ablation Study}
\label{sec:ablation}

Figure~\ref{fig:ablation} examines execution and adaptation mechanisms. Panel (a) removes functions without
readaptation, measuring the adapted harness's dependence on them.
Panel (b) repeats adaptation with the same candidate budget.
\emph{Full} includes all coordination and adaptation components.

\noindent\textbf{Coordination during execution.}
\emph{W/O Task Plan} removes goal guidance; \emph{W/O Buffer Reset}
disables visual refresh of VLA history in navigation and harness
image context in manipulation. \emph{W/O Recover} removes corrective
guidance, and \emph{W/O Completion} bypasses the completion check.
\emph{W/O Context} retains the task and concise progress, testing
compact runtime memory rather than eliminating memory.

Navigation SR falls from 57.5\% to 29.2\% without completion
checks and to 33.3\% with compact memory, the two largest
navigation losses among the execution ablations. This indicates
sensitivity to progress tracking and termination decisions.
For manipulation, removing visual refresh gives the largest L1
decrease (55.8\% to 37.5\%). Removing recovery leaves L1 nearly
unchanged but lowers L2 from 38.3\% to 23.3\%; compact memory
also lowers L2 to 22.5\%, versus 53.3\% on L1. These contrasts
are consistent with a greater role for recovery and retained
context in longer workflows with dependencies between stages.
This greater sensitivity coexists with smaller overall adaptation
gains on L2 (Table~\ref{tab:manipulation_results}).
\smallskip
\noindent\textbf{Adaptation mechanisms.}
Replacing the revision graph with flat history (\emph{W/O Graph})
reduces navigation SR from 57.5\% to 42.5\% and manipulation L2
from 38.3\% to 26.7\%, despite retaining versions and rollback.
The L1 change is smaller, from 55.8\% to 50.8\%.
With versions and rollback available in both settings, these
results support organizing related revisions and behavioral
evidence to guide adaptation under the same candidate budget,
particularly on navigation and L2.

\emph{SR Select} chooses the next candidate to refine using task
success alone, excluding trajectories and behavioral assessments
from selection. The revision agent still accesses trajectories,
and failed candidates remain eligible. SR falls to 38.3\% on
navigation, 48.3\% on L1, and 25.0\% on L2. Success alone cannot
reveal whether a patch produced its predicted effect or another
failure masked a local improvement. Without this attribution and
verification evidence, selection can overlook promising candidates.
Under the same candidate budget, \emph{SR Select} performs worse,
supporting the use of behavioral evidence during candidate selection.

In \emph{Shared Context}, all agent working contexts are shared
and carried across tasks. Navigation SR falls from 57.5\% to
49.2\%, and L2 from 38.3\% to 30.0\%, while L1 changes little
(55.8\% to 55.0\%). Together with \emph{W/O Graph}, these
results support the combined design for cross-task refinement:
organized revision evidence persists, while each agent's
working context has a separate scope and lifetime.

Assigning analysis, revision, and selection to one agent
(\emph{Single Agent}) or merging observations with hypotheses
(\emph{W/O E\&H}) also lowers navigation SR. Effects on L1 are
smaller: both \emph{W/O Recover} and \emph{W/O E\&H} reach
56.7\%, compared with 55.8\% for Full. These close means over
three runs do not establish a reliable ordering; component
benefits vary by task and its coordination demands.

\vspace{-2.2mm}
\subsection{Execution Stability}
\label{sec:execution_stability}

The common role of the harness is visible across both robot
platforms in Fig.~\ref{fig:trajectory}: it maintains task-level
progress while the VLA supplies robot-specific actions.
The simulated navigation and manipulation traces show more
goal-directed progression after adaptation, complementing the
held-out SR gains with examples of coordinated execution.

In real-world deployment, the robot deviates at R59
while approaching Goal 2 after completing Goal 1. Corrective
guidance persists through R67--R72 until Goal 2 is reached at
R75, after which execution continues without restarting. The
harness preserves the pending goal while the VLA executes the
corrective motion; the harness source remains unchanged
throughout the episode. These examples illustrate how
embodiment-specific policies can sustain task progress and
recovery under the same harness architecture.

\vspace{-2.2mm}
\subsection{Limitations and Future Work}
\label{sec:limitations}

Gains vary with the adaptation model and task demands, so the
value of individual coordination mechanisms may differ across
settings. Errors in visual or spatial reasoning may affect both
code revision and behavioral assessment~\cite{yang2025thinkingspace},
and separate contexts do not rule out errors shared by the base
model. Human-guided adaptation could provide occasional
corrections to disambiguate evidence and review hypotheses,
building on language feedback in robot learning~\cite{shi2024yell}.
Corrections supported by subsequent rollouts could then enter
adaptation memory to guide future revisions.

The present framework edits coordination code while leaving VLA
parameters unchanged. Future work could use persistent adaptation
memory to identify recurring failures and select validated
observation--action sequences for VLA fine-tuning or
reinforcement-based post-training~\cite{chen2025conrft}.
Memory would guide experience selection rather than serve
directly as action supervision. This could extend experience-driven
adaptation from coordination policies to VLA parameters.

\section{Conclusion}
\label{sec:conclusion}

We presented AdaHVLA, an adaptive harness that refines code-based
coordination policies through robot experience to better align
agent reasoning and memory with VLA execution. Its multiagent
process separates evidence analysis, harness revision, and
behavioral assessment, connecting testable coordination hypotheses
to observed robot behavior. A revision graph supports refinement
within a task, while adaptation memory carries relevant experience
into subsequent tasks and environments. Experiments on quadruped
navigation and manipulation show improved task success across
adaptation models and VLA backbones, while real-world deployment
illustrates sustained progress across task stages. These findings support coordination policy adaptation as a practical
route to more reliable long-horizon embodied execution.

\bibliographystyle{IEEEtran}
\bibliography{references}

\end{document}